\documentclass[10pt,twocolumn,letterpaper]{article}

\usepackage{iccv}              

\definecolor{iccvblue}{rgb}{0.21,0.49,0.74}
\usepackage[pagebackref,breaklinks,colorlinks,allcolors=iccvblue]{hyperref}
\usepackage{booktabs}
\usepackage{tabularx}
\usepackage{flushend}
\usepackage{xcolor}
\usepackage[accsupp]{axessibility}  
\def\paperID{*****} 
\def\confName{ICCV}
\def\confYear{2025}

\title{Automated Assessment of Aesthetic Outcomes in Facial Plastic Surgery}

\author{Pegah Varghaei\\
Michigan State University\\
{\tt\small varghaei@msu.edu}
\and
Kiran Abraham-Aggarwal\\
Cornell University\\
{\tt\small ka372@cornell.edu}
\and
 Manoj T. Abraham\\
 New York Medical College\\
{\tt\small DrAbraham@fprls.com}
\and
Arun Ross\\
Michigan State University\\
{\tt\small rossarun@msu.edu}
}

\begin{document}
\maketitle
\begin{abstract}
We introduce a scalable, interpretable computer-vision framework for quantifying aesthetic outcomes of facial plastic surgery using frontal photographs. Our pipeline leverages automated landmark detection, geometric facial symmetry computation, deep-learning-based age estimation, and nasal morphology analysis. To perform this study, we first assemble the largest curated dataset of paired pre- and post-operative facial images to date, encompassing 7,160 photographs from 1,259 patients. This dataset includes a dedicated rhinoplasty-only subset consisting of 732 images from 366 patients, 96.2\% of whom showed improvement in at least one of the three nasal measurements with statistically significant group-level change. 
Among these patients, the greatest statistically significant improvements (p \textless 0.001) occurred in the alar width to face width ratio (77.0\%), nose length to face height ratio (41.5\%), and alar width to intercanthal ratio (39.3\%). Among the broader frontal-view cohort, comprising 989 rigorously filtered subjects, 71.3\% exhibited significant enhancements in global facial symmetry or perceived age (p \textless 0.01). Importantly, our analysis shows that patient identity remains consistent post-operatively, with True Match Rates of 99.5\% and 99.6\% at a False Match Rate of 0.01\% for the rhinoplasty-specific and general patient cohorts, respectively. Additionally, we analyze inter-practitioner variability in improvement rates.
By providing reproducible, quantitative benchmarks and a novel dataset, our pipeline facilitates data-driven surgical planning, patient counseling, and objective outcome evaluation across practices.
\end{abstract}
    
\section{Introduction}
\label{sec:intro}

Facial attractiveness is a powerful driver of human perception, influencing social dynamics, professional outcomes, psychological well-being, and perceived youthfulness \cite{rhodes1998facial, little2011facial}. In clinical aesthetics, it remains one of the most sought-after yet subjective goals.

To address facial concerns, a range of cosmetic procedures—both surgical and non-surgical—are widely used. These include rhinoplasty, facelift, blepharoplasty (eyelid surgery), jaw contouring, dermal fillers, and botulinum toxin \cite{bashour2006history, farkas1994anthropometry}.  However, outcomes are often evaluated qualitatively, with limited use of reproducible, quantitative frameworks \cite{Ghahari2017}.

In facial plastic surgery, patient-reported outcome measures (PROMs) such as Rhinoplasty Outcomes Evaluation (ROE), Facelift Outcomes Evaluation (FOE), and FACE-Q are widely used to assess outcomes from the patient’s perspective \cite{alsarraf2001measuring, kosowski2009systematic, klassen2016development, klassen2015face, klassen2010measuring, pusic2013development, berger2019assessing, klassen2017face}. These instruments are subjective questionnaires in which patients rate their satisfaction with aesthetic and functional results, providing quick and procedure-specific insight into surgical success. They are valued for capturing patient-perceived benefits and quality-of-life changes that might not be evident to clinicians, offering a convenient quantitative snapshot of patient satisfaction.
However, they have limitation in objectivity. Zojaji et al. found no strong correlation between objective facial proportion changes and ROE scores, highlighting that subjective satisfaction can differ from measured facial proportions \cite{zojaji2019association}. While ROE and FOE are quick and procedure-specific, their development relied mainly on expert opinion with minimal patient input, which raises concerns about content validity \cite{zojaji2019association}. Even the more comprehensive FACE-Q, despite rigorous validation, ultimately relies on patient self-report. These considerations highlight the need for complementary objective assessment tools to provide a more complete and unbiased evaluation of surgical outcomes. Indeed, quantitative measures should mirror patient- and surgeon-reported metrics: Yilmaz et al. found postoperative gains in NOSE, ROE, and RHINO questionnaire scores \cite{ozturk2025evaluation}, and Kemal et al. observed marked improvements on patient-rated visual analog scales for nasal function and aesthetic appearance \cite{kemal2022impact}.

Attractiveness is shaped by a complex interplay of visual factors—including symmetry, proportionality, perceived youthfulness, and regional aesthetic norms—which have been extensively studied in both psychology and computer vision \cite{rhodes2001attractiveness, little2011facial}. These elements are challenging to quantify, particularly in uncontrolled clinical imagery where lighting, pose, and expression vary substantially across cases.

\begin{figure}[ht]
\centering
\includegraphics[width=0.55\linewidth]{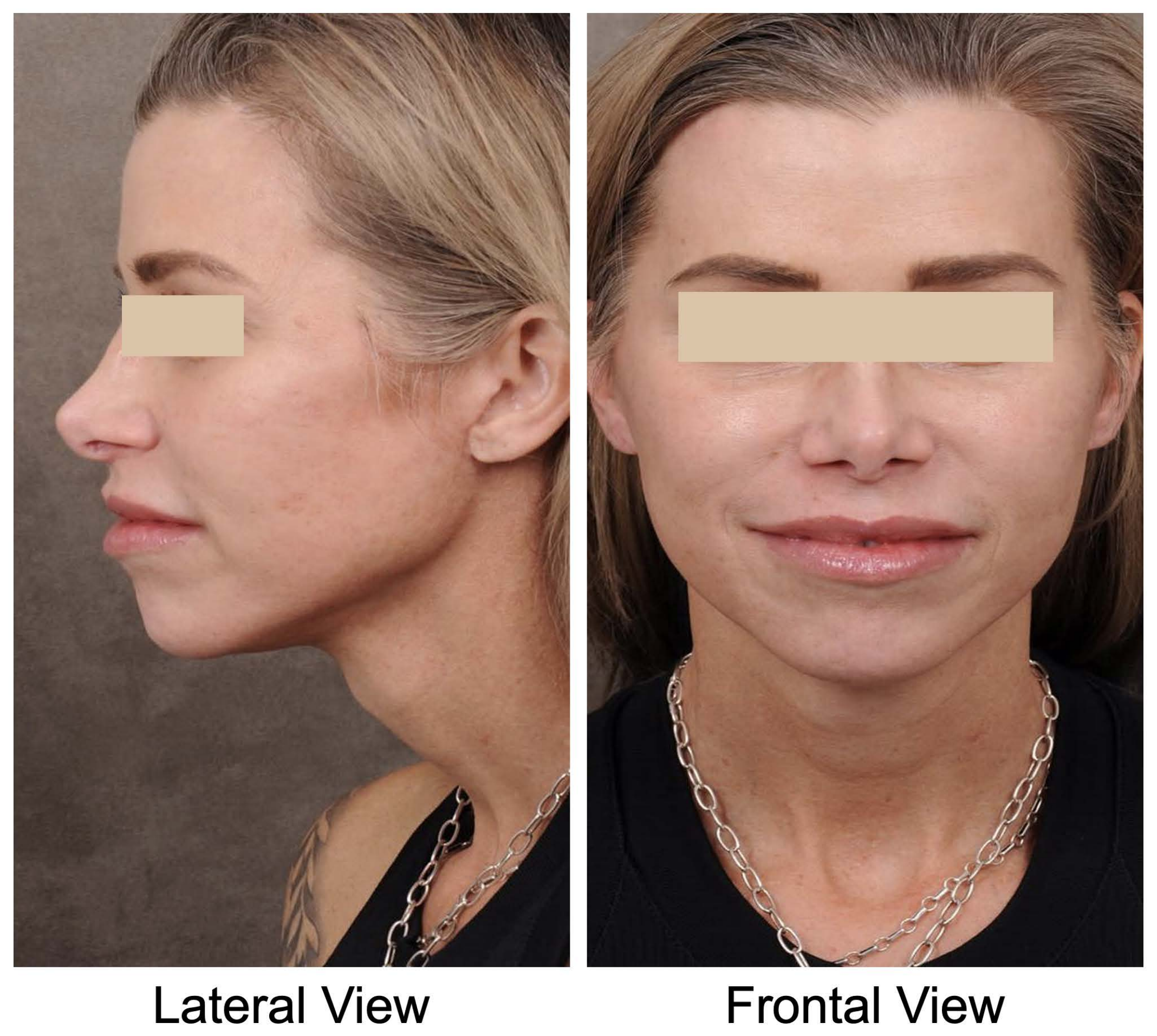}
\caption{Lateral (left) and frontal (right) facial views used for rhinoplasty outcome analysis.}

\label{fig:frontal_vs_profile}
\end{figure}

Among facial cosmetic procedures, rhinoplasty holds particular clinical and perceptual significance due to the central location and visual dominance of the nose \cite{bashour2006history}. Its outcomes can dramatically reshape perceived facial balance. However, while side-profile changes such as dorsal hump reduction and tip rotation often define surgical success, {\em automated analysis}  of lateral views is limited by inconsistent pose, occlusion, and landmarking errors \cite{borsting2020applied}. Frontal facial images, by contrast, offer a more standardized and reproducible canvas for landmark-based geometric analysis \cite{Celiktutan2013} (Figure~\ref{fig:frontal_vs_profile}). 


We develop a scalable computer-vision framework for quantitatively evaluating facial cosmetic outcomes—focusing on rhinoplasty but applicable to other aesthetic interventions—by integrating automated facial landmark detection, geometric feature analysis, and deep-learning-based age estimation on frontal pre- and post-operative images. To facilitate this, we assembled a large-scale, IRB-approved dataset of paired images from the web, uniquely annotated by procedure type. Our analyses encompass focused morphometric assessments for rhinoplasty outcomes and broader evaluations of facial symmetry and perceived rejuvenation.

To our knowledge, this is the largest real-world dataset of facial plastic surgery outcomes with paired imagery and inferred procedural labels. By combining scale, automation, and interpretability, our framework bridges the gap between subjective visual assessment and reproducible computational evaluation. It supports outcome benchmarking, and surgical planning.

\subsection*{Key Findings}

\begin{itemize}

    \item \textbf{Dataset novelty and scale:}  
    This work introduces the largest annotated dataset of facial plastic surgeries to date, using automated caption parsing to infer procedures and paired frontal imagery from real-world clinical cases.

    \item \textbf{Post-operative identity preservation:}
    Face recognition performance remained nearly perfect after surgery, with True Match Rates of 99.5\% (rhinoplasty-only) and 99.6\% (general cohort) at a False Match Rate of 0.01\%, confirming that patient identity is preserved.
    
    \item \textbf{Global facial enhancement in broader cohort:}  
    Among 989 patients undergoing mixed facial procedures, 71.3\% exhibited measurable improvements in facial symmetry, apparent age, or both.
    
    \item \textbf{Rhinoplasty-specific improvement:}  In the rhinoplasty-only cohort (366 subjects), 96.2\% of patients showed improvement in at least one of the three nasal measurements with statistically significant group-level change. Improvement was most pronounced in the alar width to face width ratio (77.0\%), followed by the nose length to face height ratio (41.5\%) and alar width to intercanthal ratio (39.3\%).

   \item \textbf{Variation across surgeons:}  
    Improvement rates ranged from 11.1\% to 65\% across practitioners, reflecting heterogeneity in technique, baseline patient characteristics, and aesthetic goals.
   
   \item \textbf{Non-improved cases:} Only 14 of 366 subjects (3.8\%) did not meet the improvement criterion on any of the three statistically significant nasal ratios (Alar Width / Intercanthal Distance, Alar Width / Face Width, Nose Length / Face Height). In this small subset, changes were mostly neutral;  Alar Width / Intercanthal Distance and Tip Midline Deviation more often worsened. 
\end{itemize}

To our knowledge, this is the first fully automated, interpretable, and scalable pipeline to quantify post-operative facial change from standardized 2D frontal photographs \cite{borsting2020applied,lo2021automatic}. By producing continuous, human-understandable metrics—global symmetry, nasal morphometry, and apparent age—our approach enables practitioner benchmarking, personalized surgical planning, and deeper insights into aesthetic outcomes across procedural contexts.

\section{Related Work}

\subsection{Facial Surgery Datasets}

Several datasets have been curated to study the effects of facial surgery on biometric recognition, spanning both 2D photographic and 3D/volumetric modalities:

\begin{itemize}
  \item \textbf{IIITD Plastic Surgery Face Database (2010)}:  
    1,800 full‐frontal pre‐ and post‐surgery images of 900 individuals, captured under controlled illumination and neutral expression. Covers 519 local (e.g., rhinoplasty, blepharoplasty) and 381 global (e.g., facelifts, skin peeling) procedures \cite{singh2010plastic}.

  \item \textbf{HDA Plastic Surgery Database (2012)}:  
    Photographic pairs across five procedure types: eyebrow correction (128), eyelid correction (131), facelift (98), nasal correction (174), and facial bone work (107), enabling per‐procedure benchmarking \cite{rathgeb2020plastic}.

  \item \textbf{LUCY CT Shape–Skeleton Facial Dataset (2015)}:  
    144 high‐resolution maxillofacial CT scans (0.48×0.48×1 mm³) from 72 patients, each scanned before and after orthognathic surgery, facilitating volumetric‐change analysis \cite{rathgeb2020plastic}.
\end{itemize}

These datasets have primarily supported research on evaluating biometric resilience in the face of surgical alteration (i.e., determining whether the identity changes after surgery), rather than aesthetic outcomes \cite{singh2010plastic,rathgeb2020plastic}; for example, Jillela and Ross \cite{jillela2012mitigating} demonstrated that face recognition models suffer significant drops in accuracy after surgery and proposed region-based fusion strategies to mitigate this loss.

Despite the existence of large “in-the-wild” face corpora, these surgical datasets remain the primary resources for analyzing facial structure changes. However, they are constrained by limited scale, labeled demographic diversity, and real-world variability, motivating the need for broader, high-volume datasets such as the one introduced in this study, which spans different surgeons, practices, and likely diverse patient populations, though formal demographic annotations are not available.

\subsection{Facial Attractiveness Metrics}
Facial attractiveness, though inherently subjective, is often associated with geometric regularities such as bilateral symmetry, proportionality, and averageness \cite{little2011facial,rhodes2001attractiveness}. These attributes form the basis of computational aesthetic assessments, particularly in the context of facial plastic surgery.

Several anthropometric ratios have been proposed to quantify these ideals, including alar width to intercanthal distance, nose length to face height, and the so-called “golden ratio” \cite{bashour2006history,farkas1994anthropometry} (Figure~\ref{fig:frontal_vs_profile}). The golden ratio (approximately 1.618) refers to a proportion considered aesthetically pleasing and has historically been used to guide ideal vertical facial thirds and horizontal fifths \cite{mantelakis2018proportions}. These metrics are frequently employed in clinical practice and underlie landmark-based evaluations of nasal proportion, facial symmetry, and overall facial harmony.

Recent work has emphasized the impact of facial beautification—whether through plastic surgery, makeup, or digital retouching—on both the perceived attractiveness and the biometric integrity of facial images \cite{rathgeb2019impact}. These modifications often align facial proportions more closely with aesthetic ideals but can introduce geometric distortions that complicate post-operative assessment or automated analysis.

Related studies have shown that similar geometric ratios are also predictive of identity and demographic traits such as gender, suggesting their discriminative power extends beyond aesthetics \cite{cao2011can}. This overlap reinforces the value of using well-established facial proportions in evaluating post-surgical outcomes \cite{hwang2021divine}.

\subsection{Landmark-Based Aesthetic Assessment}
Facial landmarks serve as pivotal reference points, enabling reproducible and precise geometric measurements necessary for aesthetic analysis. These approaches are particularly valuable due to their interpretability, alignment with classical facial proportion theories \cite{farkas1994anthropometry}, and compatibility with both traditional and deep learning models. The robustness of primary landmarks (nose tip, eye corners, mouth corners) has been well-established, particularly for pose normalization and symmetry analysis \cite{Celiktutan2013}. Ghahari et al. further demonstrate the relationship between facial landmark geometries and age perception, suggesting their applicability to aesthetic evaluations \cite{Ghahari2017}.
Recent advancements expand landmark applications for facial beauty assessment. Wang et al. introduced methods integrating texture and landmark geometry to improve age and beauty predictions, achieving superior performance compared to landmark-only methods \cite{wang2021age}. Similarly, Zhang et al. combined deep facial landmark detection with geometric morphometric techniques, demonstrating improved robustness and precision in aesthetic assessments \cite{zyang2023automated,de2022analysis}. In our study, we focus exclusively on landmark-based geometric analysis due to its higher interpretability and consistency across varied photographic conditions, avoiding potential confounds introduced by texture variations (e.g., lighting, makeup, skin quality).

\subsection{State-of-the-Art Landmark Detection}
Accurate facial landmark detection is foundational to computational aesthetics. Methods like MediaPipe FaceMesh, capable of detecting 468 facial landmarks \cite{mediapipe2023}, and high-resolution networks (HRNet) that offer high accuracy in landmark localization \cite{bianco2018aesthetics}, have transformed automated aesthetic analysis \cite{sun2019deep,jakhete2024comprehensive,liao2025humanaesexpert}. Additionally, lightweight approaches such as PFLD (Practical Facial Landmark Detector) deliver efficient yet accurate landmark detection, essential for real-time applications \cite{guo2019pfld}.

\subsection{Computer Vision for Rhinoplasty Outcome Analysis}
Automated evaluation of rhinoplasty outcomes using computer vision is a growing area of interest, albeit with limited literature. Borsting et al. introduced RhinoNet, a convolutional neural network for distinguishing between  pre- and post-operative rhinoplasty images \cite{borsting2020applied}. GAN-based approaches by Knoedler et al. demonstrate potential in simulating post-surgical outcomes for preoperative planning \cite{knoedler2024turn}. A systematic review by Han et al. further underscores the potential of AI in improving rhinoplasty outcome prediction, surgical planning, and patient satisfaction assessments \cite{eldaly2022simulation}. Machine learning frameworks like AutoGluon have been used to objectively predict outcome satisfaction from pre- and post-operative measurements \cite{topsakal2023utilization}. Stepánek et. al. applied both tree‑based classifiers and regression models to paired pre‑ and post‑rhinoplasty photographs to derive continuous, interpretable scores of facial attractiveness change, demonstrating the value of quantitative aesthetic metrics \cite{vstvepanek2018evaluation,vstvepanek2019evaluation}.

However, the aforementioned methods largely focus on classification, simulation, or black-box satisfaction prediction, without providing continuous, interpretable metrics of anatomical change. In particular, there is a lack of end-to-end frameworks that operate on standard 2D frontal photographs to quantify symmetry, nasal morphometry, and apparent age—key indicators used in clinical practice. Our work fills this gap by proposing a fully automated, interpretable pipeline that produces human-understandable measurements. 

\subsection{Challenges in Aesthetic Surgery Data}
\begin{figure}[htbp]
    \centering
    \includegraphics[width=0.45\textwidth,height=0.23\textheight]{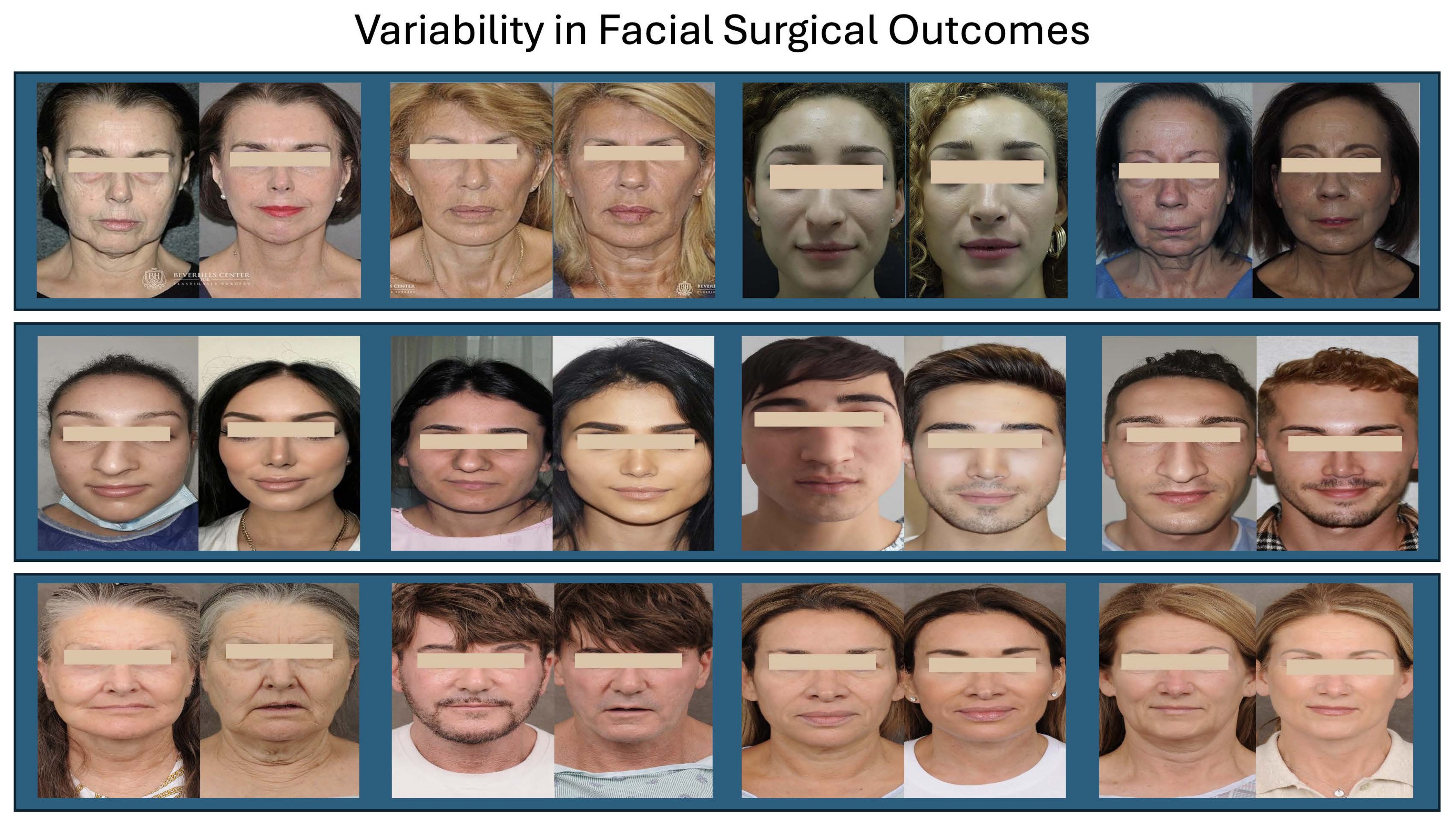}
    \caption{Paired pre- and post-op images highlighting challenges in automated aesthetic analysis, including lighting, makeup/editing, and multiple surgical procedures.}
    \label{fig:variability_outcomes}
\end{figure}

Several studies emphasize inherent challenges when working with aesthetic surgical images, including variability due to lighting conditions, makeup, regional anatomical differences, and concurrent surgical interventions \cite{Henderson2005,rhee2017simple}. As illustrated in Figure~\ref{fig:variability_outcomes}, various factors can significantly affect the visual assessment of surgical outcomes.
These challenges often complicate standardized outcome evaluations. Addressing these complexities requires targeted methodological strategies, such as restricting analysis to specific procedures like rhinoplasty and standardizing frontal facial views for robust landmark detection and consistent outcome metrics, as employed in this study.

\section{Proposed Method}

Our pipeline comprises three stages: (1) image collection and dataset assembly, (2) identity validation using face recognition, and (3) aesthetic evaluation using landmark-based analysis of facial symmetry, age, and nasal morphology. Each stage is outlined below with corresponding dataset details and analytic techniques.

\subsection{Data Collection}

For this research, we constructed a dataset of pre- and post-operative facial images collected from public clinical cases shared by surgeons on Instagram. The data collection process was approved by University Institutional Review Board (IRB), which permitted the ethical use of publicly posted images for non-commercial research purposes.

We used the Apify Instagram scraping \cite{apify_instagram_scraper} tool to systematically download high-resolution before-and-after surgery photos. Surgeons were selected based on the consistency of their posting style, favoring those who presented vertically aligned facial comparisons across frontal and lateral views. Here, posting style refers to the standardized presentation of images, including consistent angles, head positioning, and framing, which facilitates reliable visual analysis. The resulting dataset comprises images originally uploaded by clinicians based in the United States, Turkey, Italy, Iran, and several countries in South America.
We restrict our analysis to strictly frontal, quality-filtered images to ensure reliable automated landmark detection and to minimize occlusion and pose variability.  
Images with variable head pitch, missing frontal views, or failed automated landmark detection were excluded; only subjects with paired frontal pre- and post-operative images were analyzed.

\textbf{Demographic Metadata Note:} While demographic analysis and fairness auditing are important in clinical AI, our dataset does not include subject-level demographic annotations such as age, sex, or ethnicity. The only available metadata are the geographic region of the practitioner and the type of surgery performed. As a result, we are unable to perform demographic subgroup evaluation in the present study.

Due to IRB constraints and patient privacy considerations, the full dataset cannot be publicly released. However, the data processing code and annotation scripts are available here:  \href{https://github.com/varghaei/face-aesthetics-analysis}{https://github.com/varghaei/face-aesthetics-analysis}.

\begin{figure}[h]
\centering
\includegraphics[width=0.46\textwidth]{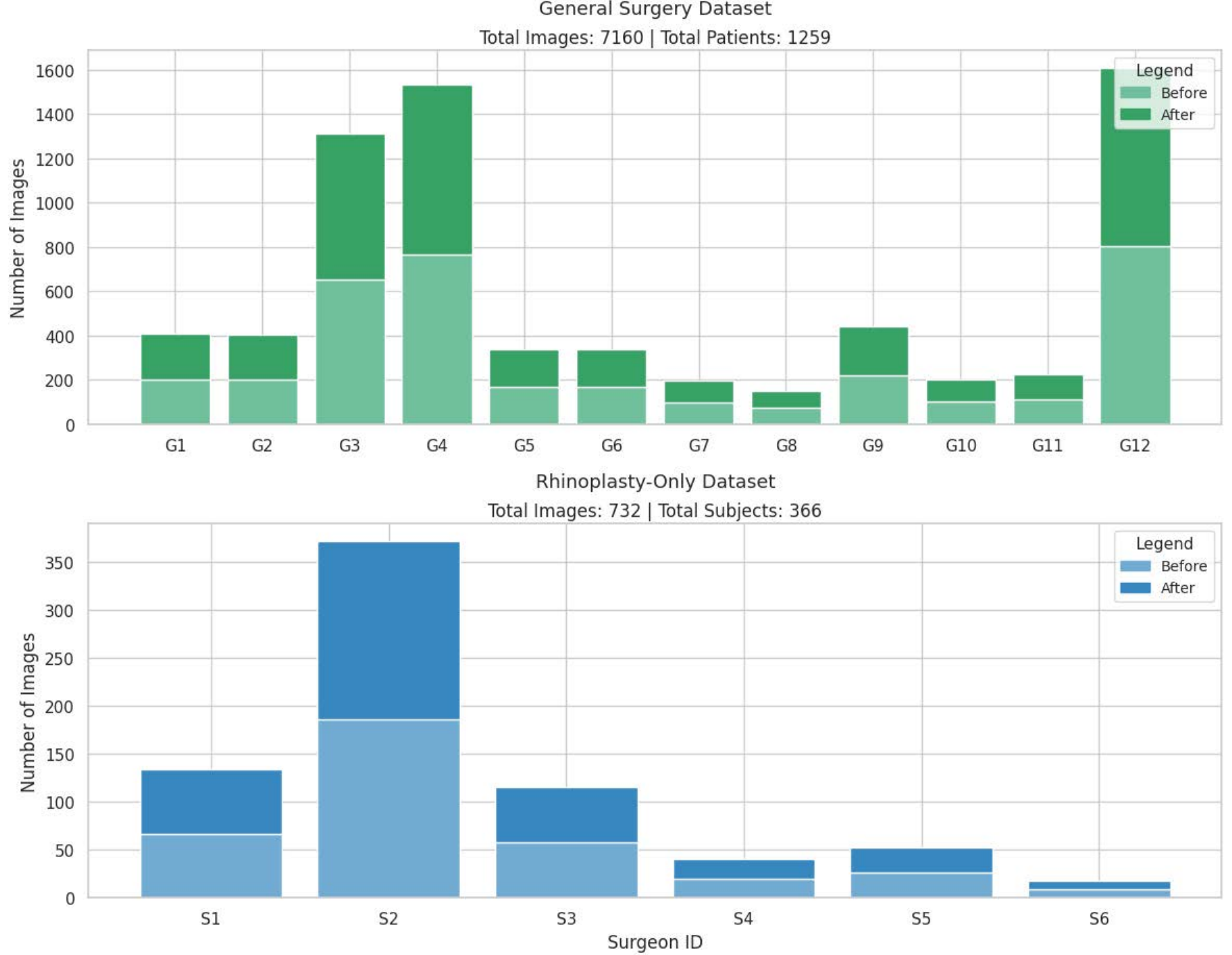}
\caption{Image counts per surgeon for general (top) and rhinoplasty-only (bottom) datasets. Light colors: pre-op; dark colors: post-op. Surgeon IDs are anonymized (G1–G12, S1–S6).}
\label{fig:general_data_bar}
\end{figure}

\begin{figure}[h]
\centering
\includegraphics[width=0.45\textwidth]{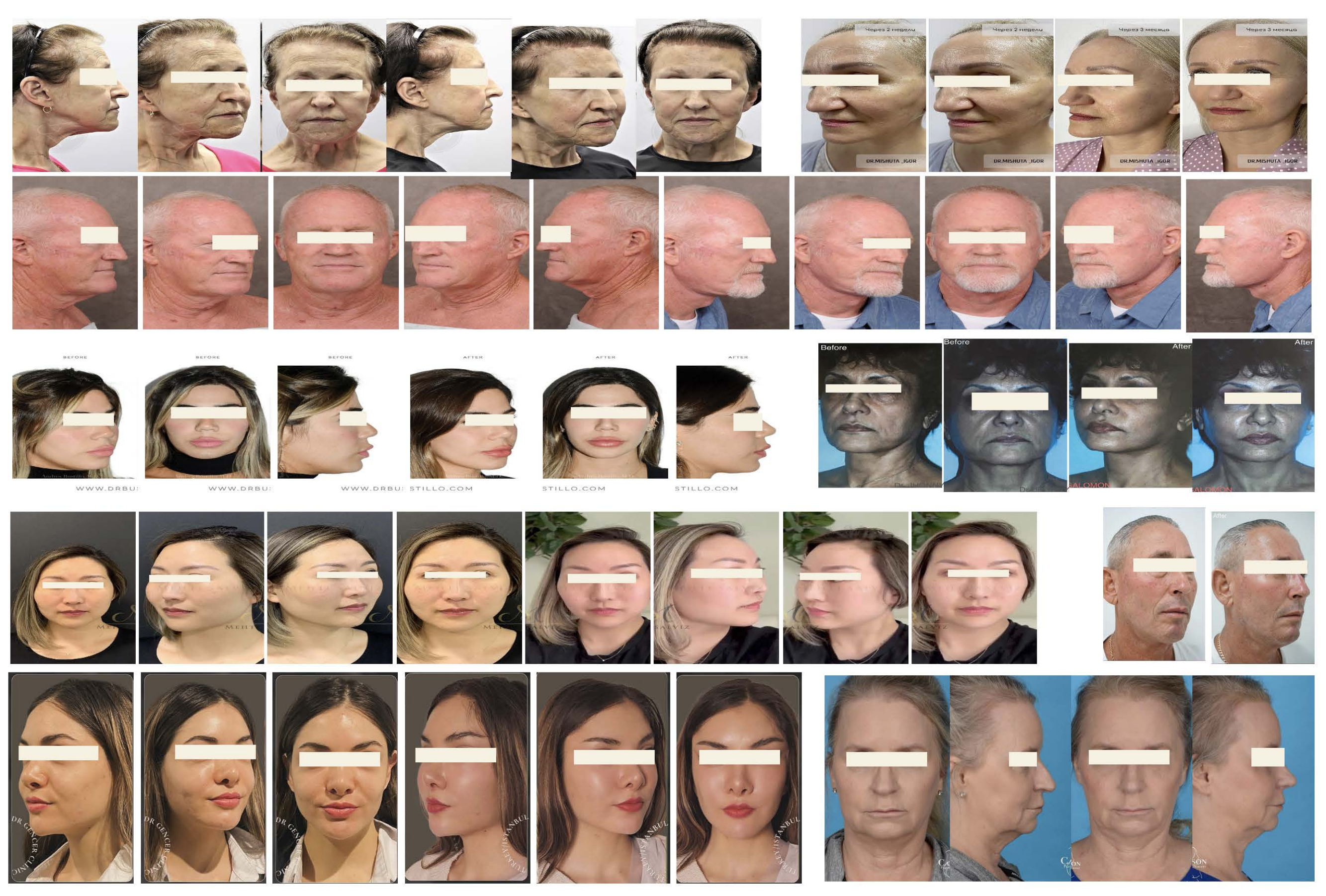}
\caption{Example pre- and post-operative images from the SurFace1259 dataset. Each row shows a representative patient from a different surgeon, highlighting variability in outcomes and photographic conditions.}
\label{fig:sample_faces}
\end{figure}

\vspace{\baselineskip}
\textbf{General Facial Surgery Dataset:} We constructed a dataset of 1,259 patients, which we refer to as \textit{SurFace1259}, who underwent various facial aesthetic procedures, including rhinoplasty, facelift, eyelid surgery, lip augmentation, and others. It includes 7,160 images (3,580 pre- and post-operative pairs) collected from 12 surgeons' public Instagram cases. Among these, 1,099 patients contributed multiple before–after pairs, allowing analysis of follow-ups or additional interventions. Despite variability in documentation, the dataset enables paired and statistical evaluation. Figure~\ref{fig:general_data_bar} summarizes image counts per surgeon, and Figure~\ref{fig:sample_faces} shows example outcomes.

\textbf{Rhinoplasty-Only Dataset:} To specifically study nasal outcomes, we assembled a separate rhinoplasty-only dataset comprising 366 patients, each contributing a paired set of pre- and post-operative frontal images (732 images total), from six surgeons whose case volumes range from 18 to 372 images. The per-surgeon image counts for this cohort are shown in the bottom panel of Figure~\ref{fig:general_data_bar}. By focusing exclusively on single-procedure rhinoplasty cases, this dataset enables targeted geometric and landmark-based analyses of nasal symmetry, proportion and alignment without confounding from concurrent procedures.

\subsection{Biometric Recognition}
To assess whether surgical changes compromise biometric identity, we evaluated pre- and post-operative facial images using AdaFace,\footnote{https://github.com/mk-minchul/AdaFace} a state-of-the-art deep face recognition model known for its robustness to facial variations \cite{kim2022adaface}. For each individual, facial embeddings were extracted from both pre- and post-surgery images. We then measured similarity between embeddings using cosine similarity.
Performance was quantified using True Match Rate (TMR) at a stringent False Match Rate (FMR) of 0.01\%. In our evaluation protocol, genuine comparisons (same identity, pre- and post-operative) were contrasted against a large number of imposter comparisons (different identities). This approach quantifies identity preservation despite aesthetic changes.

\subsection{Global Symmetry}

Facial symmetry and youthfulness are widely recognized as central components of aesthetic perception, supported by literature in psychology and computer vision \cite{rhodes1998facial,little2011facial}. To objectively assess changes in these attributes, we developed an automated analysis pipeline to compute global symmetry scores and estimate apparent age from paired pre- and post-operative facial images.

\textbf{Data Selection and Preprocessing:} From our initial pool of 1,259 pre- and post-operative subject pairs, we aimed to isolate cases suitable for consistent landmark-based facial analysis. MediaPipe FaceMesh \cite{mediapipe2023} was first applied to each image to extract facial landmarks and assess face orientation. To focus on subjects with high-quality, frontal facial views, we then manually reviewed the dataset and removed the subjects  that did not have frontal angles. After this filtering process, 989 subject pairs remained. These qualified cases were then grouped by surgeon and processed through our standardized analysis pipeline.

\textbf{Landmark Detection and Alignment:} Facial landmarks were extracted using MediaPipe FaceMesh \cite{mediapipe2023}, which produces 468 2D keypoints per face. To minimize the effects of pose variation, each image was aligned using an affine transformation derived from the coordinates of the outer eye corners (landmarks 33 and 263). This transformation normalized both scale and in-plane rotation across image pairs.

\begin{figure}[h]
\centering
\includegraphics[width=0.95\linewidth]{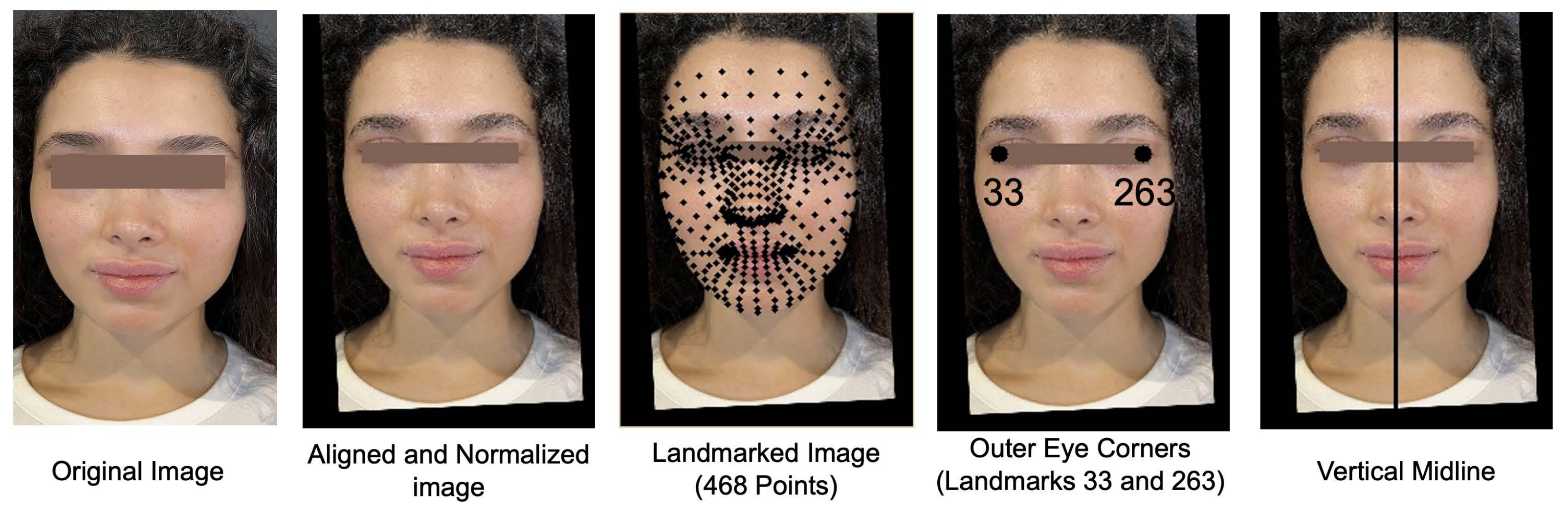}
\caption{Illustration of facial alignment and symmetry computation. Top row: original image, aligned image, and overlaid 468 facial landmarks. Bottom row: landmarks 33 and 263 (outer eye corners) used for alignment; vertical midline computed as the midpoint between them.}
\label{fig:landmark_alignment_vis}
\end{figure}

Figure~\ref{fig:landmark_alignment_vis} illustrates the full pipeline: the top row shows the original image (left), the aligned and normalized output (middle), and the overlaid set of 468 facial landmarks (right). The bottom-left panel highlights the two outer eye corner landmarks (33 and 263) used to compute the affine transformation. Finally, the bottom-right panel shows the computed vertical midline, defined as the midpoint between these two landmarks. This visual sequence clarifies the role of landmark-based alignment and midline computation in standardizing facial orientation.

\textbf{Symmetry Computation:} To quantify facial symmetry, we first define the vertical midline as the horizontal midpoint between the two outer eye‐corner landmarks (indices 33 and 263), illustrated in the bottom‐right panel of Figure~\ref{fig:landmark_alignment_vis}.
We then compute a symmetry error score based on how well landmarks on the left side align with their reflected counterparts on the right side. The procedure is as follows:

\begin{itemize}
  \item Extract all 468 facial landmarks 
    \(\{(x_i, y_i)\}_{i=1}^{468}.\)
  \item Compute the vertical midline:
    \[
      x_{\mathrm{mid}} = \frac{x_{33} + x_{263}}{2}.
    \]
  \item Partition landmarks into left and right sets:
   \[
  \begin{aligned}
    \mathcal{L}_{\mathrm{left}}
      &= \{(x_i,y_i)\mid x_i < x_{\mathrm{mid}}\},\\
    \mathcal{L}_{\mathrm{right}}
      &= \{(x_j,y_j)\mid x_j > x_{\mathrm{mid}}\}.
  \end{aligned}
\]
  \item Reflect each left‐side landmark \((x_i,y_i)\) across the midline to get
    \[
      (\tilde{x}_i, y_i)
      = \bigl(2x_{\mathrm{mid}} - x_i,\;y_i\bigr).
    \]
  \item Build a KDTree \cite{bentley1975multidimensional} on the right‐side landmarks \(\mathcal{L}_{\mathrm{right}}\).
  \item For each reflected point \((\tilde{x}_i,y_i)\), query the KDTree to find its nearest neighbor \((x'_i,y'_i)\in\mathcal{L}_{\mathrm{right}}\).
  \item Compute the symmetry distance for each pair as
    \[
      d_i
      = \sqrt{\bigl(x_i - (2x_{\mathrm{mid}} - x'_i)\bigr)^2 
             + (y_i - y'_i)^2}\,,
    \]
    i.e.,\ reflect the matched right‐side point back across the midline and measure its Euclidean distance to the original left landmark.
\end{itemize}

Finally, the overall symmetry error \(S\) is defined as the mean of these distances:
\[
S = \frac{1}{|\mathcal{L}_{\mathrm{left}}|}\;
    \sum_{(x_i,y_i)\,\in\,\mathcal{L}_{\mathrm{left}}}
    \biggl\|
      \begin{bmatrix}x_i\\y_i\end{bmatrix}
      -
      \begin{bmatrix}2\,x_{\mathrm{mid}} - x'_i\\y'_i\end{bmatrix}
    \biggr\|_2.
\]
Lower values of \(S\) indicate greater bilateral symmetry. This formulation is robust to local facial variations and inherently handles small pose discrepancies due to the alignment step described in Section 3.2.

\textbf{Apparent Age Estimation:} To evaluate perceived youthfulness—a common goal of facial cosmetic procedures—we estimated the apparent age of each face using the DeepFace framework \cite{taigman2014deepface}. DeepFace is a deep convolutional neural network originally developed for facial recognition, and later extended to predict age, gender, and emotion attributes. It uses a 9-layer neural architecture trained on millions of images and fine-tuned for facial attribute inference. 

For each subject, we applied DeepFace’s age predictor to pre- and post-operative images and computed the difference in estimated age. A decrease in predicted age is interpreted as increased apparent youthfulness. This metric reflects model-estimated age from facial appearance and is not a measure of biological aging.

\textbf{Outcome Categorization:} Each subject was assigned to one of four outcome categories:
\begin{itemize}
\item \textbf{Both:} improvement in both facial symmetry and youthfulness,
\item \textbf{Only Symmetric:} improvement in symmetry but not age,
\item \textbf{Only Younger:} improvement in age but not symmetry,
\item \textbf{Neither:} no improvement in either dimension.
\end{itemize}

\subsection{Nose-Specific Morphometric Analysis}

To complement our global assessment of facial symmetry and apparent age, we developed a region-specific evaluation framework to quantify nasal aesthetic outcomes. This analysis focuses on a curated rhinoplasty-only dataset, comprising 366 pre- and post-operative image pairs. Each subject in this cohort underwent rhinoplasty as the sole aesthetic procedure, enabling isolated evaluation of nasal changes without interference from additional interventions. The dataset includes cases from six surgeons, with the number of images per surgeon ranging from 9 to 186, reflecting a mix of high- and low-volume surgical practices.

\textbf{Landmark Extraction and Alignment:} Facial landmarks were extracted using MediaPipe FaceMesh \cite{mediapipe2023}, which detects 468 2D keypoints per face. From these, we selected key nasal and facial reference points including the nose tip (1), left and right nostrils (98 and 327), inner eye corners (133 and 362), cheeks (234 and 454), chin (152), forehead (10), and glabella (168). To standardize facial orientation and reduce pose-related variability, each image was rotated based on the angle between the inner eye corners and resized with padding to a fixed $512 \times 512$ resolution. This preprocessing step ensured consistent alignment and scale across the dataset.

\begin{figure}[h]
\centering
\includegraphics[width=0.36\textwidth]{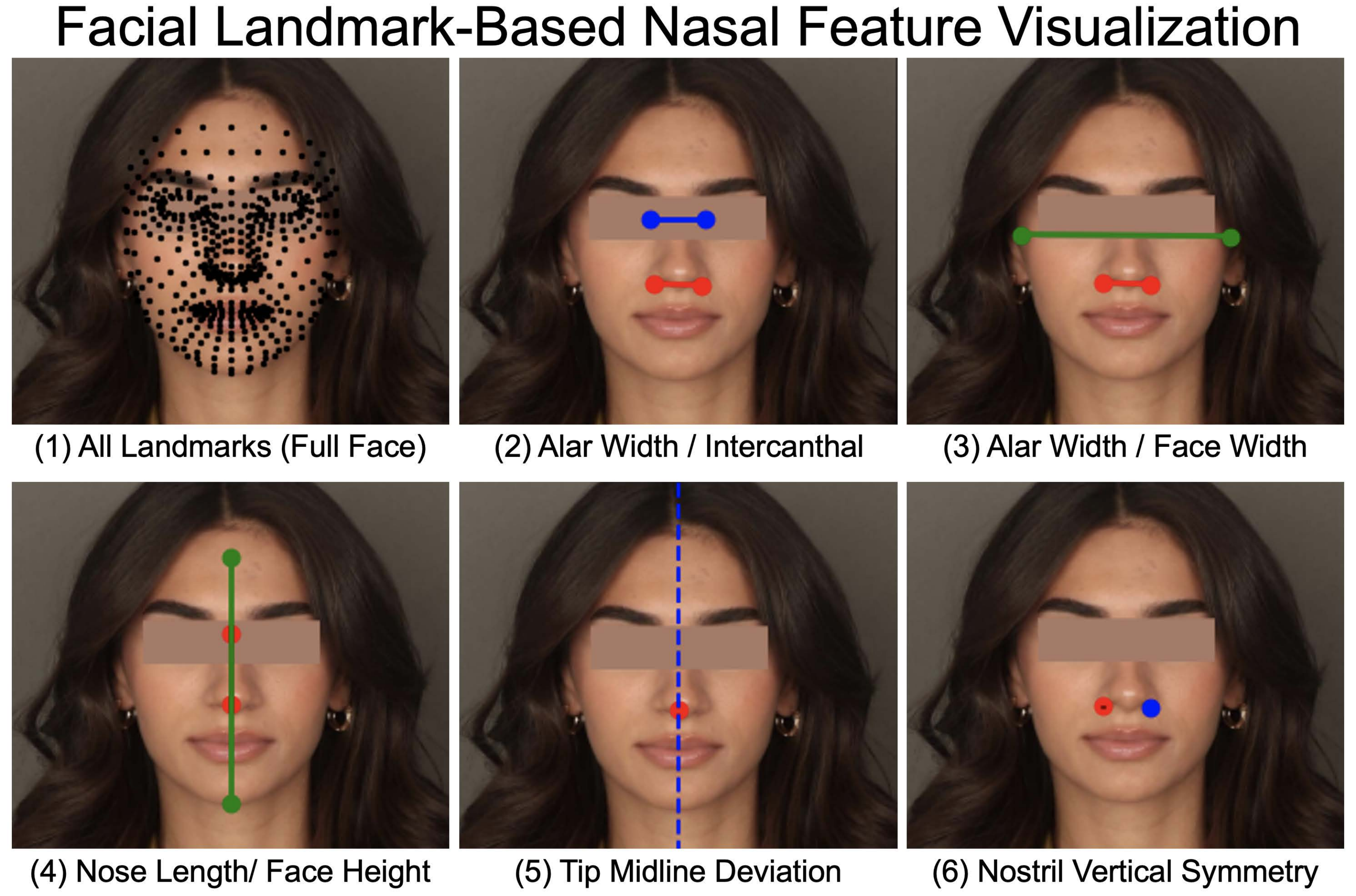}
\caption{Visualization of landmark-based nasal metrics: (1) all detected facial landmarks; (2) alar width/intercanthal distance; (3) alar width/facial width; (4) nose length/facial height; (5) tip midline deviation; (6) nostril vertical symmetry.}
\label{fig:nose_features}
\end{figure}

\textbf{Morphometric Feature Computation:} Using the aligned facial landmarks, we computed five geometric features designed to capture nasal proportion and symmetry (Figure~\ref{fig:nose_features}). These included: (1) the ratio of alar width to intercanthal distance, reflecting nasal width relative to eye spacing; (2) the ratio of alar width to full facial width, contextualizing nose size within the broader facial structure; (3) nose length divided by facial height, representing vertical nasal proportion; (4) tip midline deviation, defined as the absolute horizontal distance from the nose tip to the vertical midline of the face; and (5) nostril vertical asymmetry, calculated as the vertical height difference between the left and right nostrils.

These features were selected based on established literature in facial anthropometry and aesthetic plastic surgery \cite{farkas1994anthropometry,bashour2006history,pallett2010new}, which emphasize proportionality, midline alignment, and bilateral symmetry as critical indicators of nasal attractiveness. The chosen metrics align with common clinical evaluation practices, enabling interpretable and reproducible comparisons across patients and practitioners.

\textbf{Rhinoplasty Cohort Metrics:} We processed all 366 rhinoplasty-only image pairs through the described morphometric pipeline. For statistical reporting, a subject was labeled “improved” if any one of the three nasal features with statistically significant group-level change (alar width to face width ratio, nose length to face height ratio, alar width to intercanthal ratio) shifted closer to its anthropometric ideal. Given that symmetry, alignment, and proportion are established aesthetic goals in rhinoplasty, improvement in even a single frontal-view feature is considered sufficient to indicate a positive surgical outcome.

\section{Results}

\begin{figure}[ht]
\centering
\includegraphics[width=0.46\textwidth]{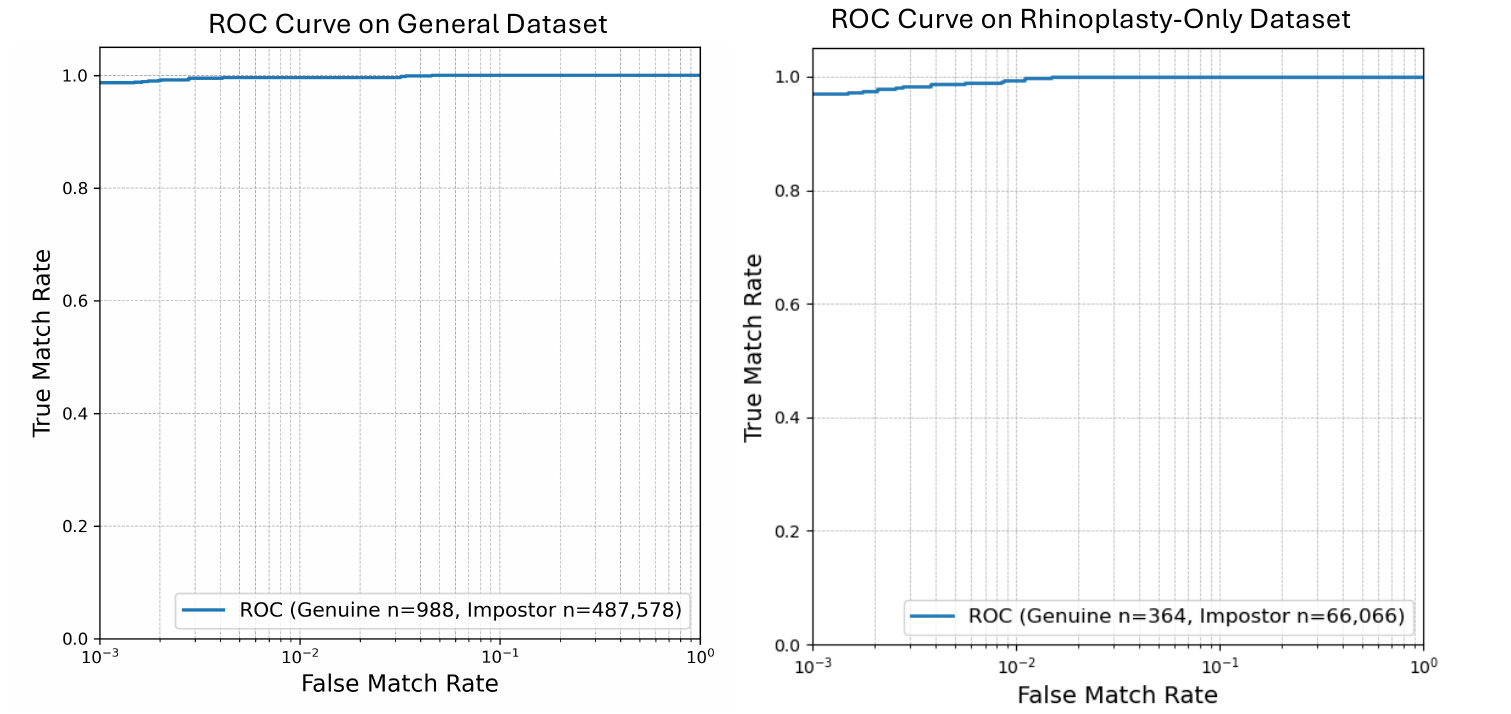} 
\caption{
ROC curves using AdaFace features. \textbf{Left:} General dataset. \textbf{Right:} Rhinoplasty-only dataset.
}
\label{fig:roc_curves}
\end{figure}

\textbf{Biometric Identity Preservation:} Recognition remained near-perfect across both cohorts: 99.5\% TMR for the rhinoplasty-only group and 99.6\% for the general dataset, at an FMR of 0.01\%. These results suggest that while visible aesthetic changes occur post-operatively, biometric identity
is strongly preserved. Figure~\ref{fig:roc_curves} shows ROC curves for both datasets, confirming minimal recognition degradation post-surgery.

\begin{figure}[ht]
\centering
\includegraphics[width=0.42\textwidth]{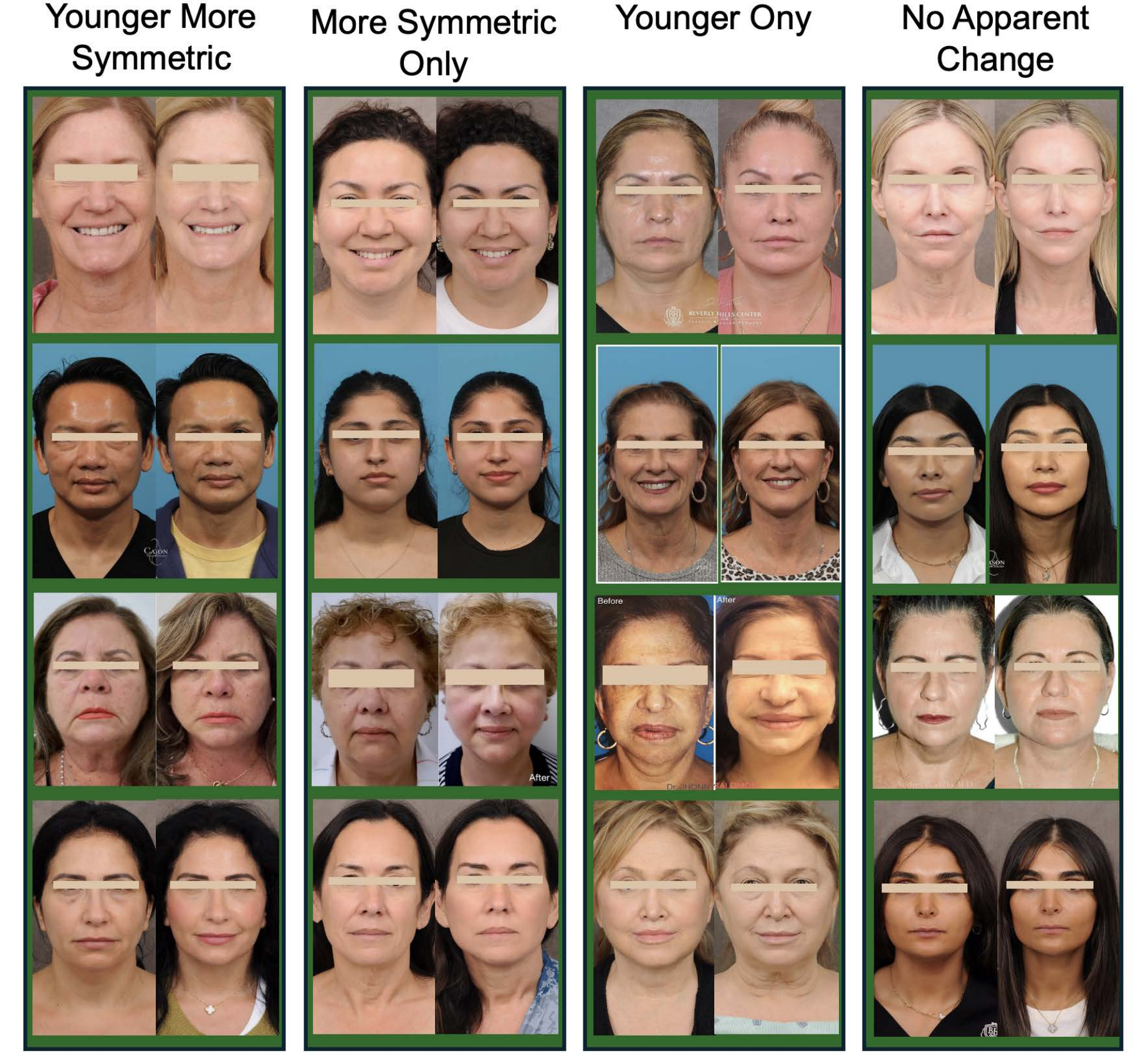}
\caption{Examples of pre- and post-operative outcomes: Younger \& More Symmetric, Symmetric Only, Younger Only, No Change.}
\label{fig:qual_symm}
\end{figure}

\textbf{Global Symmetry and Apparent Age Outcomes:} Among 989 evaluated cases, 22.1\% improved in both symmetry and apparent age, 28.4\% in symmetry only, 20.7\% in age only, and 28.7\% showed no improvement, yielding 71.3\% with measurable aesthetic gains. Figure~\ref{fig:qual_symm} shows visual examples of each outcome category.

Compared to prior work (e.g., a 4.3-year AI-estimated reduction post-facelift \cite{zhang2021turning}), our 42.8\% age improvement rate confirms significant perceived youthfulness gains across a broader set of procedures. Our dataset lacks pre-to-post timing information, limiting assessment of long-term outcomes. We plan to incorporate time metadata in the future to better separate transient effects (e.g., swelling) from lasting improvements.

\begin{figure}[ht]
\centering
\includegraphics[width=0.9\columnwidth]{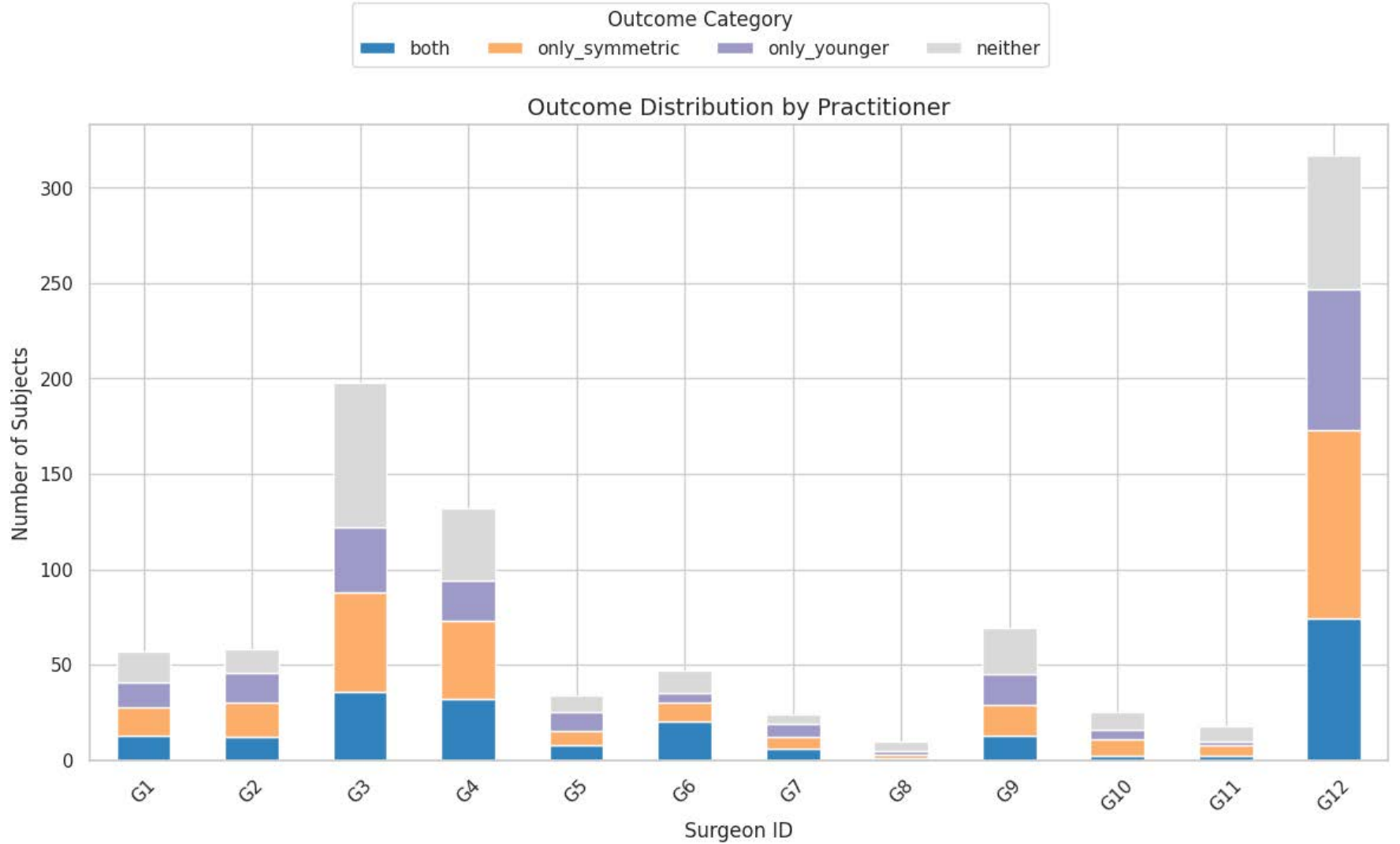}
\caption{Aesthetic outcome distributions by surgeon: improvements in both dimensions (Both), only symmetry, only youthfulness, or neither.}
\label{fig:doctor-results}
\end{figure}

\textbf{Practitioner-Level Outcomes:} As shown in Figure~\ref{fig:doctor-results}, practitioner outcomes varied: some had more dual improvements, others mixed or neutral. Most showed a higher share of positive outcomes than neutral or negative ones, potentially reflecting differences in technique or case selection. We lacked patient-level demographic data (e.g., age, sex, ethnicity), so we were unable to analyze how these factors affect outcomes.

\subsection{Nose Morphometric Outcomes}

\textbf{Feature-Level Results:} Among the 366 rhinoplasty‐only cases, statistically significant improvements ($p < 0.001$) were observed in 77.0\% for Alar Width/Face Width (AW/FW), 41.5\% for Nose Length/Face Height (NL/FH), and 39.3\% for Alar Width/Intercanthal Distance (AW/IC). Improvements in Tip Midline Deviation (41.3\%) and Nostril Vertical Symmetry (27.0\%) were not statistically significant ($p > 0.05$). Overall, 96.2\% of cases showed improvement in at least one of the three statistically significant nasal measurements (45.9\% in exactly one, 38.8\% in exactly two, and 11.5\% in all three), demonstrating that the vast majority achieved a measurable refinement in their primary nasal concern. These findings align with classical anthropometric ideals \cite{farkas1994anthropometry}, providing the first large-scale quantitative assessment of distinct nasal refinement outcomes in facial plastic surgery.

\begin{figure}[htbp]
  \centering
\includegraphics[width=0.35\textwidth]{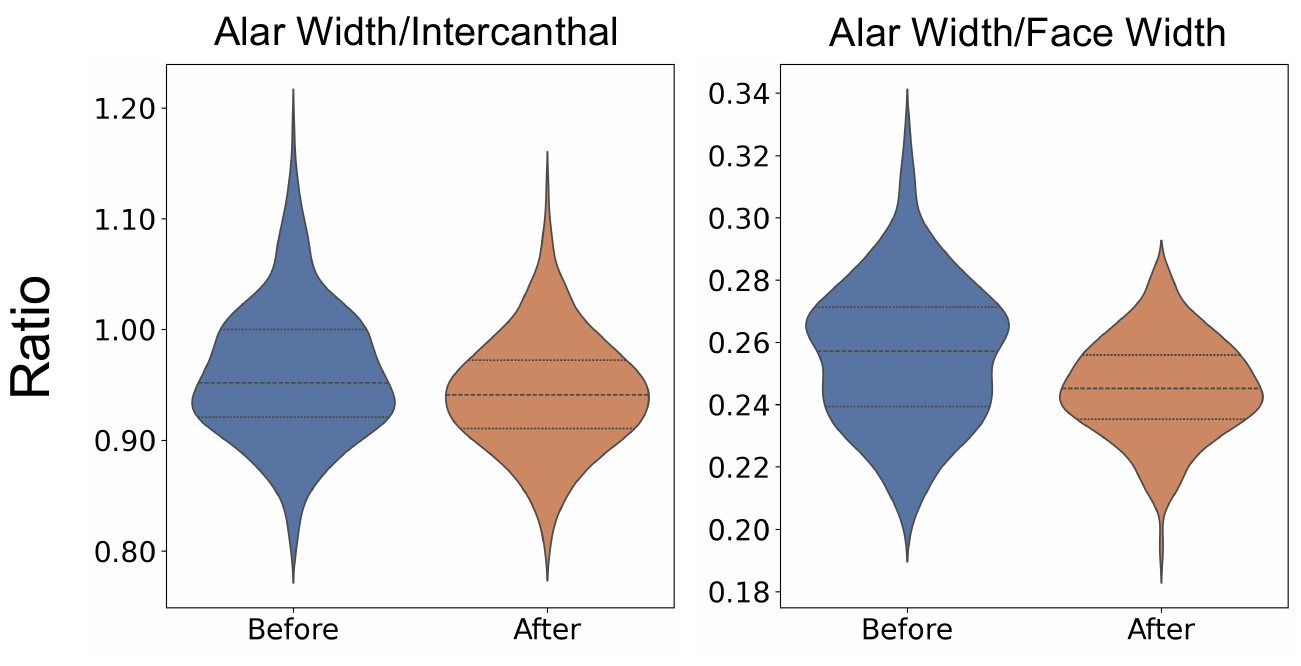}
  \caption{Distribution of alar ratios pre- and post-rhinoplasty for 366 patients: (left) Alar/Intercanthal; (right) Alar/Face Width.}
  \label{fig:alar_ratio_distributions}
\end{figure}

\textbf{Discrepancy Between Alar Ratios:} Although facial width and intercanthal distance remain unchanged post-surgery, we define improvement as moving closer to the ideal ratio:
\[
|\;R_{\text{after}} - R_{\text{ideal}}\;| \;<\;|\;R_{\text{before}} - R_{\text{ideal}}\;|,
\]
where, \(R_{\text{ideal}}=1.0\) for Alar/Intercanthal and \(0.20\) for Alar/Face Width. Since most patients start below 1.0 but above 0.20, the same narrowing moves them away from one ideal but closer to the other—explaining the higher improvement rate for Alar/Face Width (see Figure~\ref{fig:alar_ratio_distributions}).

\textbf{Practitioner Comparison:} Improvement rates were uniformly high: three surgeons achieved 100\% overall, and the remainder were in between 92.3\% and 94.8\%. For the highest-volume surgeon, the AW/FW ratio improved in roughly 95\% of cases.

\textbf{Analysis of Non-Improved Cases:} For the three significant nasal ratios AW/FW, NL/FH, and AW/IC, 352 of 366 cases (96.2\%) improved in more than one metric; 14 cases (3.8\%) improved in none, termed “non-improved.” This definition excludes non-significant endpoints such as tip deviation. The absence of post-operative time metadata limits our ability to differentiate temporary changes from long-lasting surgical outcomes.

\section{Conclusion and Future Work}

We proposed an automated, interpretable pipeline for assessing facial cosmetic outcomes from paired frontal images. Across 1,259 cases, 71.3\% showed improvement in either symmetry or youthfulness. In the rhinoplasty-only subset (n=366), Alar Width/Face Width improved in 77.0\% of patients. Biometric identity was observed to be preserved, with TMR of 99.5\% in the rhinoplasty only group and 99.6\% in the general dataset at an FMR of 0.01\%.

Since images were sourced from public web content, some may have been digitally retouched, potentially biasing morphometric results. Still, our landmark-based and deep-learning metrics offer a scalable alternative to subjective assessment.

This study is limited to 2D imagery and does not include 3D geometry, which may help improve the aesthetic assessment accuracy. We also note that models like AdaFace and DeepFace introduce some opacity, though their outputs align with our interpretable pipeline. Independent dataset validation is needed to test generalizability. The lack of demographic metadata also precludes subgroup analysis.

Future work will integrate profile views, surgical notes, and end-to-end models linking facial changes to patient satisfaction. We also aim to add time-since-surgery metadata to better distinguish lasting results from transient postoperative effects.

\section*{Ethics Statement}
This study was approved by the Michigan State University IRB (STUDY00011708). Data were sourced from publicly shared social media posts by medical professionals. To protect patient privacy, all faces in figures have been de-identified and private metadata excluded. There was no commercial use or model training using this data. Our goal is to promote ethical AI in aesthetic medicine, with attention to consent, fairness, and transparency.

{

    \small
\bibliographystyle{ieeenat_fullname}
    \bibliography{main}
}

\end{document}